\UseRawInputEncoding
\documentclass[sigconf]{acmart}
\usepackage{graphicx}
\usepackage{amsmath}

\usepackage{subfig}
\usepackage{booktabs}
\usepackage{multirow}
\usepackage{mathtools}
\usepackage[linesnumbered,ruled]{algorithm2e}
\usepackage{float}
\usepackage{color}
\usepackage{ulem}
\usepackage[capitalise]{cleveref}
\usepackage[aboveskip=1pt,belowskip=-6pt]{caption}
\usepackage{threeparttable}
\AtBeginDocument{%
  \providecommand\BibTeX{{%
    \normalfont B\kern-0.5em{\scshape i\kern-0.25em b}\kern-0.8em\TeX}}}

\copyrightyear{2024} 
\acmYear{2024} 
\setcopyright{acmlicensed}\acmConference[ICMR '24]{Proceedings of the 2024
	International Conference on Multimedia Retrieval}{June 10--14, 2024}{Phuket,
	Thailand}
\acmBooktitle{Proceedings of the 2024 International Conference on Multimedia
	Retrieval (ICMR '24), June 10--14, 2024, Phuket, Thailand}
\acmDOI{10.1145/3652583.3658081}
\acmISBN{979-8-4007-0619-6/24/06}

\begin{document}

\title{Deep Image Clustering Based on Curriculum Learning and Density Information}

\author{Haiyang Zheng}
\orcid{0000-0001-8733-9696}
\affiliation{%
	\institution{Harbin Institute of Technology, Shenzhen}
	\city{Shenzhen}
	\country{China}
}
\email{haiyangzheng1122@gmail.com}

\author{Ruilin Zhang}
\orcid{0000-0002-4818-9282}
\affiliation{%
	\institution{Harbin Institute of Technology, Shenzhen}
	\city{Shenzhen}
	\country{China}
}
\email{zzurlz@163.com}

\author{Hongpeng Wang}
\orcid{0000-0001-8108-2674}
\authornote{indicates corresponding author.}
\affiliation{%
	\institution{Harbin Institute of Technology, Shenzhen \& Peng Cheng Laboratory}
	\city{Shenzhen}
	\country{China}
}
\email{wanghp@hit.edu.cn}

\renewcommand{\shortauthors}{Haiyang Zheng, Ruilin Zhang, \& Hongpeng Wang}

\begin{abstract}

Image clustering is one of the crucial techniques in multimedia analytics and knowledge discovery. Recently, the Deep clustering method (DC), characterized by its ability to perform feature learning and cluster assignment jointly, surpasses the performance of traditional ones on image data. However, existing methods rarely consider the role of model learning strategies in improving the robustness and performance of clustering complex image data. Furthermore, most approaches rely solely on point-to-point distances to cluster centers for partitioning the latent representations, resulting in error accumulation throughout the iterative process. In this paper, we propose a robust image clustering method (IDCL) which, to our knowledge for the first time, introduces a model training strategy using density information into image clustering. Specifically, we design a curriculum learning scheme grounded in the density information of input data, with a more reasonable learning pace. Moreover, we employ the density core rather than the individual cluster center to guide the cluster assignment. Finally, extensive comparisons with state-of-the-art clustering approaches on benchmark datasets demonstrate the superiority of the proposed method, including robustness, rapid convergence, and flexibility in terms of data scale, number of clusters, and image context.

\end{abstract}

\ccsdesc[300]{Computing methodologies~Computer vision}

\keywords{Deep clustering, Curriculum Learning, Density information, Clustering assignment, Learning pace
}



\maketitle

\section{Introduction}\label{Introduction}

In recent years, the volume of multimedia data has grown at a phenomenal rate. Also, most multimedia data are typically raw, i.e., the label information is missing or incomplete. Analyzing massive multimedia data such as images under unsupervised context to discover hidden knowledge is a challenging task. Image clustering, as a key tool in multimedia machine learning, aims to group the images by ensuring that images within the same group exhibit high homogeneity while different groups possess maximum dissimilarity. Recently, approaches that incorporate deep learning (referred to as deep clustering, DC) have demonstrated remarkable performance for image data. Technically, deep clustering seeks to transform the target data from the original space into an embedded space via neural networks, thereby concurrently executing feature learning and cluster assignment. In this way, existing deep clustering methods prefer to focus on the methodological level, including stacking complex modules\cite{VaDE, LGCC}, adding auxiliary learning objectives\cite{DTC, IDEC, DEC-DA, SCDCC, DipDECK, DeepDPM}, and improving supervised signals, \cite{EDESC, ICDM, IDECF, DEMC}. It should be noted that few attempts have been made to explore the model learning strategy during optimization, such as learning in a meaningful or rational order. However, it is significant for improving the performance and efficiency of the target model, especially for unsupervised tasks.

Interestingly, human learning offers valuable insights for optimizing neural network models. One prominent insight is the principle of progressing from simple to complex tasks, accompanied by gradual refinement. In view of this, Bengio et al. \cite{CL} propose a curriculum-based training strategy named Curriculum Learning (CL), which aims at training a model from easier data to harder data. While a few deep clustering methods incorporating Curriculum Learning have been proposed, they typically employ a loss-based posterior approach to score samples, with gradually increased size and difficulty of the training set. In fact, this learning strategy might not be optimal for clustering tasks, as model optimization under unsupervised conditions lacks ground truth labels and inevitably incorporates subjective loss design. Consequently, the strong correlation (positive correlation) often observed between loss values and sample difficulty in supervised tasks may be compromised in this context.

Furthermore, existing deep clustering typically inherits the ``minimum distance towards cluster centers'' to achieve cluster assignment in the embedding space. This behavior, derived from partitioning-clustering, might be suboptimal for image data as it relies on a limited number of cluster centers and solely on single distance information, resulting in unstable clustering outcomes or supervised signals for training.

%

In this paper, we introduce inherent density information from the data into the entire process of deep clustering and propose an image clustering method (IDCL), which is capable of handling image data with challenging scenarios such as multiple-cluster, small-scale, large-scale, and complex backgrounds. The main contributions can be summarized as follows:

1) We propose a robust network learning scheme for image clustering. Rather than relying on external loss, the density information is first employed to objectively evaluate the difficulty of the samples for the model. Moreover, we adopt a variable pace based on geometric growth, which resembles a human learning pace, to enable model optimization through a natural incremental process. With this in mind, our model converges more rapidly and can extract salient and discriminative features from complex images.


2) We develop a robust cluster assignment method, which utilizes the dynamic density core instead of the mean-forming cluster center to represent an individual cluster jointly, to enhance clustering results while refining the supervisory signal for training.

3) The quantitative experiments conducted on 10 datasets, along with in-depth analysis from four aspects (convergence, runtime, sensitivity, and ablation), effectively showcase the robustness and efficiency of our algorithm.

\section{Related Work}

\subsection{Deep Clustering}

As a pioneering work, deep embedded clustering (DEC) in \cite{DEC} is developed to jointly perform representations and cluster assignment through end-to-end joint training. Due to its promising performance and flexibility, the joint learning framework proposed by DEC is widely adopted and extended, becoming one of the most popular solutions among existing DC technologies. Currently, most studies tend to concentrate on the network backbone or feature learning, which includes stacking intricate modules\cite{VaDE, LGCC}, introducing auxiliary techniques (data augmentation, regularization, or parameter adaption)\cite{DTC, IDEC, DEC-DA, SCDCC, DipDECK, DeepDPM}, and enhancing supervised signals (i.e. enhancing pseudo-labelling reliability)\cite{EDESC, ICDM, IDECF, DEMC}.

In addition to the above efforts, several studies aim to enhance the generalization and clustering performance of deep image clustering by introducing learning strategies, such as active learning\cite{SCAL, IECAL}, contrastive learning\cite{CC}, and transfer learning\cite{TSC}. For instance, LNSCC\cite{LNSCC} employs a contrastive learning approach to exploit the positive and negative aspects of sample pairs, effectively separating clusters. However, despite their strong theoretical foundation, these works predominantly rely on auxiliary data, supervisory information, or additional learning objectives, thereby constraining the practical application of clustering.

More importantly, we observe that existing deep clustering techniques primarily rely on distance to cluster centers during the cluster assignment, which easily yields undesirable supervised signals and consequently suffers from error accumulation throughout the alternating stages of representation learning and clustering. With this in mind, we leverage the inherent density information within the data to propose an enhanced cluster partitioning method.

\subsection{Clustering with Curriculum Learning}

Drawing inspiration from the structured learning sequences in human education, curriculum learning commences with simpler examples before gradually incorporating more complex instances. For instance, in learning to write, students initially focus on fundamental rules such as spelling, grammar, and tense for each word. They then progress to mastering advanced rhetorical devices, including similes, metaphors, metonymy, and parallelism. Existing curriculum learning frameworks can be classified as an amalgamation of Difficulty Measurer and Training Scheduler \cite{SurveyCL}, where the former ascertains the relative simplicity of the training data, and the Training Scheduler determines when to introduce more challenging data during training. Initially, the Difficulty Measurer arranges all training examples in ascending order of difficulty and passes them to the Training Scheduler.  As the training advances, the Training Scheduler incrementally incorporates more challenging data into the training process.

Jiang et al.'s seminal work \cite{WWDCSC} is the first to integrate curriculum learning into deep clustering models. They devise a promising solution with theoretically and analytically supported rationale, dynamically assessing the difficulty of each sample during training and alternately optimizing sample weights and model parameters. To address cluster imbalance in training data selection, ClusterGAN \cite{ClusterGAN} introduces an exclusive loss regularization term, ensuring that the chosen samples encompass all clusters. Beyond difficulty measurement and sample selection efforts, ASPC-DA \cite{ASPC-DA} proposes an innovative adaptive step size to accommodate model training, further augmenting the model's convergence rate. Similarly, the derivative work \cite{DCSPC}, based on \cite{ASPC-DA}, incorporates such training schemes into more sophisticated deep networks like CAE, enhancing clustering performance on complex image datasets. While the incorporation of curriculum learning into deep clustering algorithms has led to notable advancements in robustness and convergence speed, limitations may arise from overreliance on loss. Specifically, the aforementioned methods depend on manually defined losses to gauge sample difficulty and subsequently drive model training using either fixed or loss-based variable step sizes, introducing subjectivity. More critically, loss-based design is impractical, particularly in complex data scenarios, as unsupervised losses are less dependable than supervised ones due to the absence of ground truth guidance. In contrast, our proposed method leverages the density information inherent in the data to develop curriculum learning.

\section{Proposed Method: IDCL}
\begin{figure*}[]
	\centering
	{\includegraphics[width=0.96\linewidth]{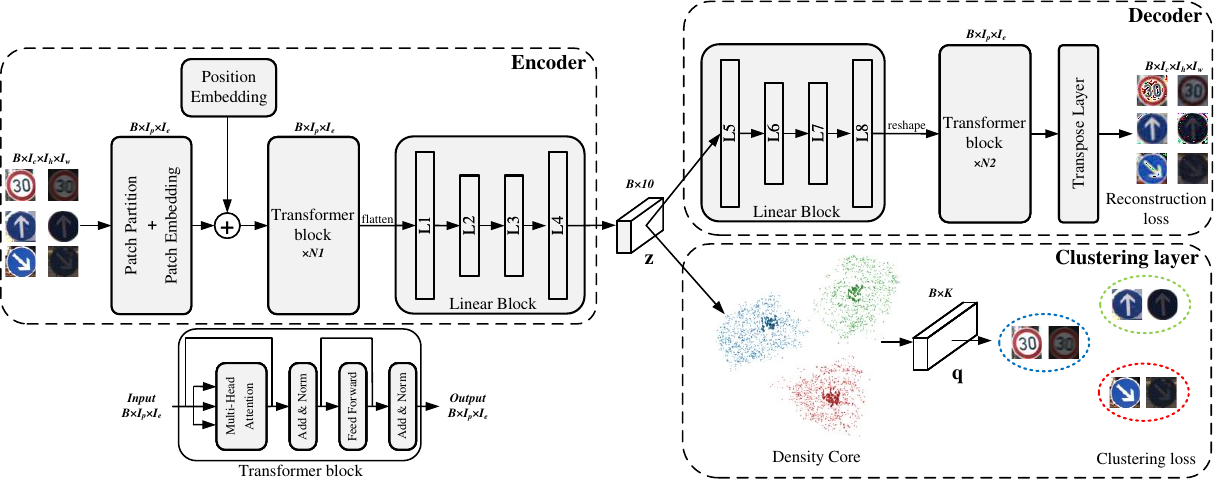}} \hspace{0mm}
	\caption{The framework of our proposed IDCL. }
	\label{Fig1}
\end{figure*}

In this research study, we aim to address the problem of partitioning dataset $X = \{x_1, x_2, \ldots, x_n\}$, containing $n$ samples, into $K$ disjoint clusters. Our model comprises a Transformer-based autoencoder (TAE) and a Clustering layer, with a detailed illustration provided in \cref{Fig1}.
The TAE consists of Encoder and Decoder networks, denoted by $f_{\mathbf{W}}(\cdot)$ and $g_{\mathbf{U}}(\cdot)$, where $\mathbf{W}$ and $\mathbf{U}$ represent the Encoder and Decoder parameters, respectively. The Encoder employs a CNN-based Patch Embedding layer to process input data of dimensions $I_c$, $I_h$, and $I_w$ into patch embeddings with dimensions $I_p$ and $I_e$, where $I_p$ denotes the number of patches and $I_e$ signifies the dimension of each patch embedding. Conventional Transformer blocks and absolute position encoding, as depicted in \cref{Fig1}, follow the design in \cite{Transformer} and are employed within the Encoder. The subsequent linear block functions analogous to the encoder network in an autoencoder. The structure of the Decoder is symmetrical to the Encoder, except for the Transpose layer, which is implemented to reassemble the patches into their original input shape. Detailed configurations of the TAE can be found in \cref{Experimental Settings}. The latent representations of $X$ in the embedded space are generated by the Encoder and denoted by $Z = \{z_1, z_2, \ldots, z_n\}$, where $z_i = f_{\mathbf{W}}(x_i) \in \mathbb{R}^d$ for $i = 1, \ldots, n$, and $d$ indicates the latent space dimension. The Decoder output for input data $x_i$ is represented by $g_{\mathbf{U}}(f_{\mathbf{W}}(x_i))$. Given the latent representations, the Clustering layer employs a robust curriculum learning-based training scheme and an enhanced cluster assignment approach to facilitate competitive clustering performance.

\subsection{Determining the Training Data with Curriculum Learning}

In clustering analysis, there exists an intuitive assumption that a cluster in the data space should be a relatively dense connection region surrounded by a low-density region. We note that the points located in the core region, owning a higher density, provide crucial information about the cluster structures, which is essential for the model to learn the actual distribution of the target data. Points distant from the core region of the actual cluster, primarily exhibiting local relative sparsity and skewness, pose challenges for the model to fit but are equally significant because they serve as the ``boundary'' against other clusters.

Based on this intuitive observation, we incorporate density information to differentiate easier (more confident) samples and subsequently add them to the training set at an earlier stage. Specifically, we define a Difficulty Measurer $\Delta$ to assign scores to all samples and a Curriculum Generator $G$, which supplies a new curriculum (i.e., an updated training set) for the model to learn in each iteration, progressively incorporating more challenging samples in a rational and incremental manner.

Given a sample $x_i \in X$, we define $\delta_i$ as the difficulty score of $x_i$. Formally, we first calculate the local density of $x_i$ in the embedded space, denoted as
\begin{equation}
\label{EQ1}
\begin{aligned}	
\rho_i = \sum_{j=1}^{n} \exp\left(-\frac{\|z_i - z_j\|^2_2}{d_c^2}\right),
\end{aligned}
\end{equation}
where $d_c$ acts as the sampling radius of density calculation, which is determined by sorting $n^2$ distances between latent representations in $Z$ from small values to large ones and selecting a number ranked at $\lambda_1$. Refer to \cite{DPC}, $\lambda_1=2\%$ is the most appropriate.
By this way, the difficulty score $\delta_i$ is formulated from MAX-MIN Normalization.
The higher the score $\delta_i$, the denser the distribution around the sample $x_i$ in embedded space, so the earlier $x_i$ should be added to the training set.
Given input data $\{x_1,x_2,\ldots,x_n\}$ and a constant $\lambda_1$. The Difficulty Measurer $\Delta$ produces difficulty scores $\{\delta_1,\delta_2,\ldots,\delta_n\}$ as outputs via the Encoder $f_{\mathbf{W}}(\cdot)$, \cref{EQ1}, as follows:
\begin{equation}
\label{EQ3}
\begin{aligned}	
\Delta(x_1, x_2, ..., x_n, \lambda_1) = \{\delta_1, \delta_2, ..., \delta_n\}.
\end{aligned}
\end{equation}

It is also instructive to note that in the human learning process, fine-grained learning of the basics helps master more challenging curricula later. Therefore, we actively set the learning pace of the model on the training set flexibly to "slow then fast" instead of "fixed" to better simulate human learning.
Let $\zeta^t \in (0,1)$ denote the proportion of training samples selected in the $t$-th epoch during model optimization. Based on the observations mentioned above, in order to achieve a "slow then fast" learning pace, $\zeta^t$ should meet the following two conditions:
\begin{equation}
\label{EQ4}
\begin{aligned}
n*\zeta^{t+1} \geq n*\zeta^{t};
\end{aligned}
\end{equation}
\begin{equation}
\label{EQ5}
\begin{aligned}
(n*\zeta^{T+1}-n*\zeta^{T}) \geq (n*\zeta^{t+1}-n*\zeta^{t}),\text{ s.t. } T>t.
\end{aligned}
\end{equation}
In particular, $n*\zeta^t$ represents the number of training samples selected in the $t$-th iteration. \cref{EQ4} illustrates the incremental inclusion of additional samples in the training set, while \cref{EQ5} indicates that the initial growth of the training set is relatively slow, with a more rapid expansion occurring in subsequent stages.

Considering the end-to-end property, the number of training samples and rate of increase (pace) in each iteration should be fed automatically by a specific continuous flow. To accomplish this, We build a unary function $F_{\zeta}$ such that $\zeta^{t}=F_{\zeta}(t)$. In this context, two types of elementary functions are available, i.e., the power ($t^{\alpha},\alpha \geq2$) and exponential ($a^{t}, a \ge 1$) functions satisfy the above conditions. Typically, the growth rate of the exponential function significantly outpaces that of the power as the independent variable increases; hence, the pacing function $F_{\zeta}$($\cdot$) in model learning is modeled in the exponential form to fast convergence with iterative training.
Formally, the pacing function $F_{\zeta}(\cdot)$ is expressed as follows.



\begin{equation}
\label{EQ6}
\begin{aligned}	
F_{\zeta}(t)=2^{\left( -\frac{\log _{2} \zeta^0}{T_{grow}} t+\log _{2} \zeta^0\right)}\text {  s.t. }T_{grow}>0, t \in N^+,
\end{aligned}
\end{equation}
where $t$ refers to the current iteration epoch, and $\zeta^0$ denotes the initial proportion of available samples, $T_{grow}$ denotes the number of iterations needed for $F_{\zeta}(\cdot)$ to attain a value of 1 for the first time. 

Mathematically, the proof that the pacing function $F_{\zeta}(\cdot)$ satisfies the aforementioned requirements is as follows.
%
%
%
%
%
%
%
%
%
\begin{proof}
	The pacing function $F_{\zeta}(\cdot)$ satisfies the "slow then fast" training requirement.
	
	
	Consider the function $F_{\zeta}(t) = 2^{\left( -\frac{\log_2 \zeta^0}{T_{grow}} t + \log_2 \zeta^0 \right)}$ with $T_{grow} > 0$ for $t \in [0, +\infty)$. We compute the first and second derivatives with respect to $t$:
	$$F_{\zeta}'(t) = 2^{-\frac{\log_{2}\zeta^0}{T_{grow}}t+\log_{2}\zeta^0}\cdot \left(-\frac{\log_{2}\zeta^0}{T_{grow}}\cdot\ln2\right),$$
	$$F_{\zeta}''(t) = 2^{-\frac{\log_{2}\zeta^0}{T_{grow}}t+\log_{2}\zeta^0}\cdot\left(\frac{\log_{2}\zeta^0}{T_{grow}}\right)^2\cdot\ln^2{2}.$$
	Observe that $F_{\zeta}'(t) > 0$ for $t \in [0, +\infty)$. This implies that $F_{\zeta}(t)$ is monotonically increasing. Thus, for $\epsilon_1 > 0$, we have $F_{\zeta}(t + \epsilon_1) > F_{\zeta}(t)$.
	
	By applying the Mean Value Theorem, there exist $\xi_1 \in (t, t + \epsilon_2)$ and $\xi_2 \in (T, T + \epsilon_2)$ such that:
	$$F_{\zeta}'(\xi_1) = \frac{F_{\zeta}(t + \epsilon_2) - F_{\zeta}(t)}{\epsilon_2},$$ 
	For $T > t$, we find that $\xi_2 > \xi_1$. As $F_{\zeta}''(t) > 0$, this implies that $F_{\zeta}'(t)$ is monotonically increasing, and thus, $F_{\zeta}'(\xi_2) > F_{\zeta}'(\xi_1)$. Therefore, if $\epsilon_2 > 0$, we obtain:$$(F_{\zeta}(T + \epsilon_2) - F_{\zeta}(T)) > (F_{\zeta}(t + \epsilon_2) - F_{\zeta}(t)).$$
	Consequently, for $\zeta^t = F_{\zeta}(t)$, we have:
	$$n \cdot \zeta^{t+\epsilon_1} \geq n \cdot \zeta^{t},\text{ s.t. } \epsilon_1>0;$$
	$$(n \cdot \zeta^{T+\epsilon_2} - n \cdot \zeta^{T}) \geq (n \cdot \zeta^{t+\epsilon_2} - n \cdot \zeta^{t}),\text{ s.t. } T\textgreater t,\epsilon_2>0$$
	Both conditions are satisfied. 
	Q.E.D.
	
\end{proof}


Note that after $T_{grow}$ epochs, we do not employ all the samples from the raw dataset for model training but remove some "extreme ones." Because we find that some samples, scoring the lowest by $\Delta$, are extreme and meaningless, such as outliers accidentally generated during real-world data acquisition, which would misguide the model and gradually deviate from the actual learning objective. 
\begin{equation}
\label{EQ7}
\begin{aligned}	
\zeta^{t}=\min \left(\zeta^{max},2^{\left( -\frac{\log _{2} \zeta^0}{T_{grow}} t+\log _{2} \zeta^0\right)}\right),
\end{aligned}
\end{equation}
where $\zeta^{max}$ denotes the maximum proportion of the sample to be selected. 

Roughly speaking, with respect to the input data $X$ at the $t$-th $(t \textgreater 1)$iteration, we can partition the samples into easy and difficult categories through the Difficulty Measurer $\Delta$ and pacing function $F_{\zeta}(\cdot)$. We define $v_i^t$ as an indicator of whether $x_i$ is an easy sample at the $t$-th $(t \textgreater 1)$iteration, as follows:
\begin{equation}
\label{EQ8}
v_{i}^{t}=\left\{\begin{array}{ll}
1 & \text{if }  x_{i} \in C_{k}, \delta^{t}_{i} \geq \mathcal A[\zeta^{t} *|C_{k}|];\\
0 & \text{ othewise, }
\end{array}\right.
\end{equation}
where $C_k$ denotes the $k$-th cluster generated from the results of the $(t-1)$-th iteration, $\delta^{t}_{i}$ denotes the difficulty score of $x_i$ in the $t$-th iteration. $\mathcal A=\text{sort}_\Downarrow(\Delta^t(C_k, \lambda_1) )$ is an ordered sequence and $\mathcal A[\zeta^{t} *|C_{k}|]$ is the value corresponding to the $\zeta^{t} *|C_{k}|$ position in $\mathcal A$. $v_i=1$ implies that $x_i$ is an easy sample; otherwise, it is classified as a difficult sample. At the $t$ iteration, the training set: 
\begin{equation}
\label{EQ9}
\begin{aligned}	
\mathcal D^t = \{x_i | x_i \in X, v_i^t=1\}.  \\
\end{aligned}
\end{equation}
It should be emphasized that the training set needs to be regenerated at the $(t+1)$ iteration as $\mathcal D^{t+1} = \{x_i | x_i \in X, v_i^{t+1}=1\}$.

Based on the aforementioned definitions and descriptions, the Curriculum Generator $G$ is used to generate a new training set during each iteration, as shown in \cref{Algorithm1}.
%

\begin{algorithm}[htp]
	\DontPrintSemicolon
	\SetAlgoLined
	\KwIn {dataset $X$; the Difficulty Measurer $\Delta$; training epoch $iter$; the proportion $\lambda_1$ of selecting the density sampling radius $d_c$.}
	\KwOut {training dataset $\mathcal D^{iter}$.}
	Compute scores for all samples using $\Delta$: $\{\delta_i\}_{i=1}^n=\Delta(X,\lambda_1)$;\\
	Determine the proportion of selected samples $\zeta^{iter}$ using \cref{EQ7}; \\
	$\mathcal D^{iter}=\emptyset$;\\
	\For{$k=1$ to $K$ }{
		Sort samples within cluster $C_k$ in descending order of difficulty scores; \\
		Compute the indicator $v^{iter}$ for each sample in cluster $C_k$ using \cref{EQ8};\\
		$\mathcal D^{iter} = \mathcal D^{iter} \cup\{x_i|x_i \in C_k,v_i^{iter}=1\}$;  \\ 	
		
	}
	\KwResult{$\mathcal D^{iter}$ }
	\caption{Curriculum Generator $G$}
	\label{Algorithm1}
\end{algorithm}

\subsection{Cluster Assignment with Density Core}

During the cluster assignment, a cluster center is typically considered to represent an individual cluster and thus quantify the distance between unclassified samples and clusters. It is worth noting that the cluster centers formed in the first-order moment way, which most of the deep clustering employs, consider only the geometric information of the clusters. From an algorithmically robust view, relying on some cluster centers might be clustering unfriendly, especially for some complex image data, as existing deep clustering adopts cluster centers that are not actual samples but only the spacial centroid. Such crisp cluster assignment of the data in an intermediate training iteration may cause undesirable supervised signals and then mislead the training procedure, which can further lead to the accumulation of errors during the alternating stages of representation learning and clustering.

For cluster $C_k$ in $X$, we describe the ``density core'' as a shrunken region consisting of a set of loosely connected points with relatively high densities in a cluster, denoted as follows

\begin{equation}
\label{EQ10}
\begin{aligned}	
\mathbf C_k^{\rm core} = \{x_i|x_i \in C_k, \rho_i \geq {\hat \rho}^k \},
\end{aligned}
\end{equation}
where ${\hat \rho}^k$ represents the value at the $\lambda_2$ position in the descending sequence of density values for all samples within cluster $C_k$ in the embedded space. Thanks to the introduction of density information, the density core formed by some high-confidence samples allows our method to abandon the overreliance on weak cluster centers while simultaneously improving the robustness and clustering accuracy. In the present study, we establish the threshold parameter $\lambda_2$ at 5\%. 

In this way, we formulate the similarity between sample $x_i$ and cluster $C_k$, called density core similarity $s_{ik}$, to replace the previous cluster-center-based one. The density core similarity $s_{ik}$ is mathematically defined as follows:
\begin{equation}
\label{EQ11}
\begin{aligned}	
s_{ik}=\sum_{x_j \in \mathbf C_k^{\rm core}} \frac{1}{(1+\|z_{i}-z_j\|^2)}
\end{aligned}
\end{equation}

\begin{figure}[]
	
	\centering
	\subfloat[Visualization of latent representations at the 50th iteration]
	{\includegraphics[width=0.97\linewidth]{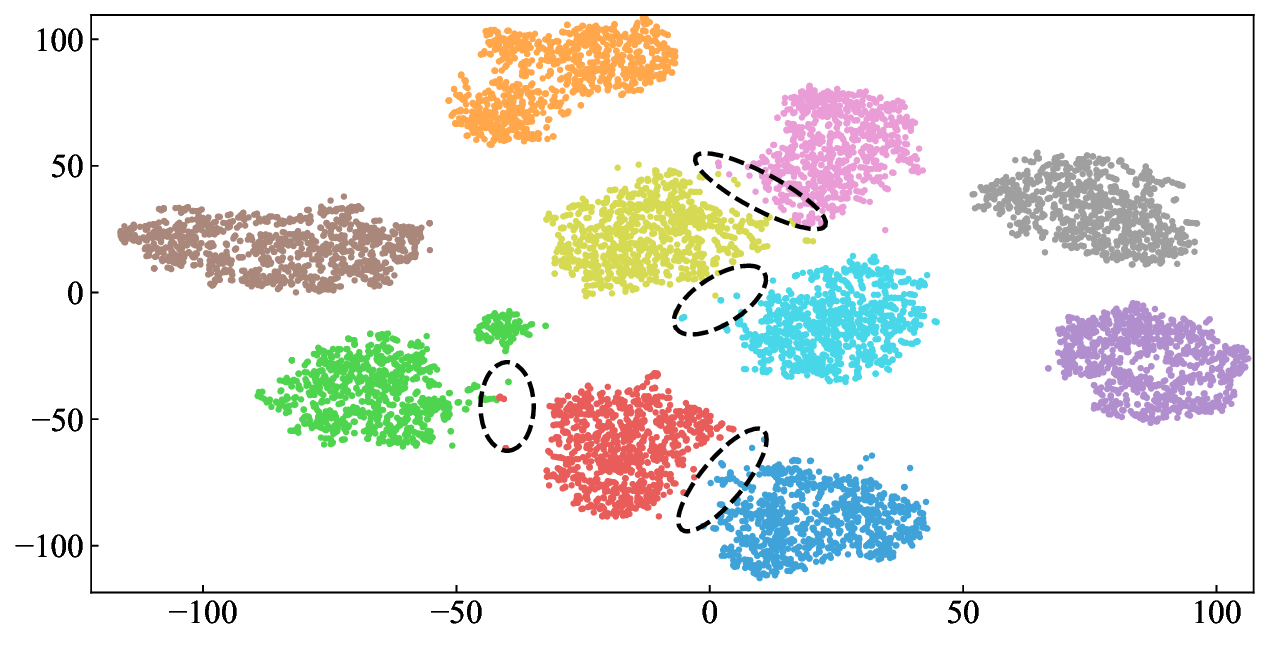}} \hspace{0mm}
	\subfloat[Assignment using cluster center]
	{\includegraphics[width=0.46\linewidth]{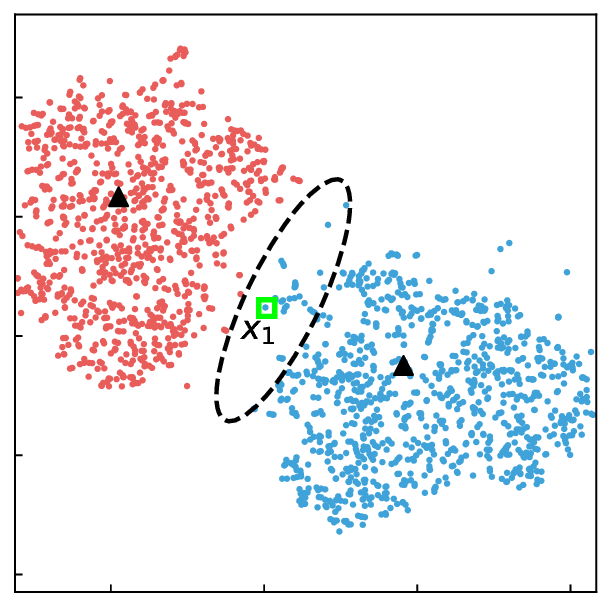}} \vspace{0mm} 
	\subfloat[Assignment using density core]
	{\includegraphics[width=0.46\linewidth]{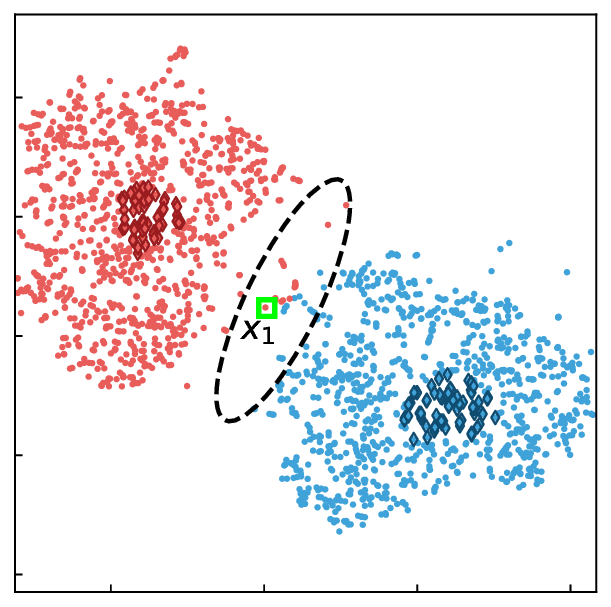}} \vspace{0mm}
	\caption{Schematic illustration of Cluster Assignment on MNIST-test dataset.}
	\label{Fig2}
\end{figure}

\cref{Fig2} demonstrates the significance of density core similarity. Specifically, \cref{Fig2}(a) is derived from the 50th iteration of the MNIST-test dataset, and the samples enclosed in dashed ellipses are difficult to be correctly partitioned by methods using cluster centers. We examine the partition of the two clusters in red and blue from \cref{Fig2}(a) as an  example. The two black triangles in \cref{Fig2}(b) signify the positions of the cluster centers, with the assumption that $C_1$ and $C_2$ denote the centers of the red and blue clusters, respectively. In this case, sample $x_1$, depicted as a green square, would be assigned to the blue cluster due to $Dist(x_1, C_1)=26.92 \textgreater Dist(x_1, C_2)=20.70$, which clearly contradicts the actual result. For our approach, as illustrated in \cref{Fig2}(c), the dark-colored points in both clusters are the identified density core. By employing density core similarity, we obtain $s_{11}=2.11 \ge s_{12}=2.34$. Thus $x_1$, along with numerous other points enclosed by the dashed ellipse, are accurately partitioned into the red cluster.

In addition, we use the student t distribution as the kernel to predict the probabilistic cluster assignment as follows.
\begin{equation}
\label{EQ12}
\begin{aligned}	
q_{ik} = \frac{\sum_{x_j \in \mathbf C_k^{\rm core}} (1+\|z_{i}-z_j\|^2)^{-1}}{\sum_{k}\sum_{x_{j} \in \mathbf C_k^{\rm core}} (1+\|z_{i}-z_{j}\|^2)^{-1}},
\end{aligned}
\end{equation}
where $q_{ik}$ indicates the probability of the $i$-th sample belonging to the $k$-th cluster and $z_i=f_{\mathbf{w}}(x_i) \in Z$.
In order to allow for unsupervised self-training, we first employ an auxiliary target variable $p_{ik}$ to refine the model predictions iteratively, as described in \cite{DEC}. 

\subsection{Optimization}
The proposed algorithm utilizes the Curriculum Generator $G$ to generate an updated training set during each iteration. Subsequently, the model is updated using backpropagation with clustering loss $L_{clu}$ and reconstruction loss $L_{rec}$ to extract latent representations of the training samples in the current training set.




The clustering loss $L_{clu}$ is defined as the Kullback -- Leibler (KL) divergence between the cluster assignment matrix $Q$ and auxiliary distribution matrix $P$. Regarding $\mathcal D^{t}$ for the $t$-th iteration, we denote the clustering loss as $L_{clu}^t$, with the formula given below.
\begin{equation}
\label{EQ14}
L_{clu}^{t}=K L(P | Q)=\sum_{i=1}^{|\mathcal D^t|} \sum_{k=1}^{K} p_{ik} \log \frac{p_{ik}}{q_{ik}},
\end{equation}
where $q_{ik}$ and $p_{ik}$ are defined according to the \cref{EQ12} and \cite{DEC}, respectively. 
The reconstruction loss $L_{rec}$ is defined as the mean squared error (MSE) loss between all samples in the training set and the objects reconstructed by the Decoder. Similarly, for $\mathcal D^{t}$, the reconstruction loss  $L_{rec}^{t}$, is defined as:
\begin{equation}
\label{EQ15}
L_{rec}^{t}=\frac{1}{|\mathcal D^t|} \sum_{i=1}^{|\mathcal D^t|}\left\|x_{i}-g_{\mathbf{U}}\left(f_{\mathbf{W}}\left(x_{i}\right)\right)\right\|_{2}^{2}.
\end{equation}
The overall loss is expressed by a weighted linear combination of $L_{clu}$ and $L_{rec}$. Specifically, for the training set $\mathcal D^{t}$, the overall loss $L^t$ is then formulated as:
\begin{equation}
\label{EQ16}
L^t = \frac{1}{|\mathcal D^t|} \sum_{i=1}^{|\mathcal D^t|}\left\|x_{i}-g_{\mathbf{U}}\left(f_{\mathbf{W}}\left(x_{i}\right)\right)\right\|_{2}^{2}+\alpha*\sum_{i=1}^{|\mathcal D^t|} \sum_{k=1}^{K} p_{i k} \log \frac{p_{i k}}{q_{i k}}
\end{equation}

By maintaining the auxiliary distribution $P$ constant during the optimization of network parameters, the gradient of $L_{clu}$ with respect to the embedded point $z_i$ can be computed as follows:
\begin{equation}
\label{EQ17}
\begin{aligned}
\frac{\partial L_{clu}}{\partial z_i} &= 2\sum_{k=1}^K (\frac{p_{ik}-q_{ik}}{\sum_{x_j \in \mathbf C_k^{\rm core}} (1+\|z_{i}-z_j\|^2)^{-1}})\\
&\quad \times (\sum_{x_j \in \mathbf C_k^{\rm core}} \frac{\|z_i-z_j\|}{(1+\|z_i-z_j\|^2)^2}),
\end{aligned}
\end{equation}
where the density core $C_k^{\rm core}$ of the $k$-th cluster is re-selected every epoch, alongside with each curriculum, as elaborated in \cref{Algorithm2}. 

If we have a mini-batch with $m$ samples and a learning rate of $\eta$, the weights of the Decoder can be updated using the formula:
\begin{equation}
\label{EQ18}
\mathbf {U}=\mathbf {U}-\frac{\eta}{m} \sum_{i=1}^{m} \frac{\partial L_{rec}}{\partial \mathbf {U}}.
\end{equation}
Here $\frac{\partial L_{rec}}{\partial \mathbf {U}}$ represents the gradient of the reconstruction loss with respect to the Decoder weights $\mathbf {U}$.

To update the weights of the Encoder, we use the formula:
\begin{equation}
\label{EQ19}
\mathbf {W}=\mathbf {W}-\frac{\eta}{m} \sum_{i=1}^{m}\left(\frac{\partial L_{rec}}{\partial \mathbf {W}}+\alpha \frac{\partial L_{clu}}{\partial \mathbf {W}}\right),
\end{equation}
where $\frac{\partial L_{rec}}{\partial \mathbf {W}}$ represents the gradient of the reconstruction loss with respect to the Encoder weights $\mathbf {W}$ and $\frac{\partial L_{clu}}{\partial \mathbf {W}}$ represents the gradient of the clustering loss with respect to the Encoder weights $\mathbf {W}$. The entire algorithm is summarized in \cref{Algorithm2}.

\begin{algorithm}[htp]
	\DontPrintSemicolon
	\SetAlgoLined
	\KwIn {Data: $X$; Number of clusters: $K$; Stopping threshold: $\mu$; Maximum iterations: $MaxIter$; Number of rounds to reach the maximum sample proportion: $T_{grow}$.}
	\KwOut {Clustering labels $Y$.}
	Initialize $\mathbf W$ and $\mathbf U$ using \cref{EQ15}; \\
	\For{iter=1 to MaxIter}{
		
		Compute all embedded points $\{z_i=f_{\mathbf{W}}(x_i)\}_{i=1}^n$; \\
		Compute density of all embedded points using \cref{EQ1}; \\
		Assign embedded points with $k$-means to obtain initial $K$ clusters $C_1,C_2,...,C_K$;  \\ 
		Update density cores $C_1^{\rm core},...,C_K^{\rm core}$ using \cref{EQ10};  \\
		Update $P$ using \cref{EQ12} and $\{z_i\}_{i=1}^n$; \\
		Save last assignment labels: $Y_{old} = Y$.\\
		Compute new assignment labels: $Y=\{\arg\max\limits_k q_{ik} \}_{i=1}^n$; \\
		\If{ $s u m(Y_{o l d}\neq Y)/n<\mu$}{  
			\textbf{End For} 
		}
		Update training set $D^{iter}$ using \cref{Algorithm1}; \\
		\For{mini batch $S $ in $D^{iter}$}{
			Compute loss $L$ of $S$ using \cref{EQ16};\\
			Update $\mathbf W$ and $\mathbf U$ using \cref{EQ18} and \cref{EQ19};\\
		}	
	}
	\KwResult{$ Y$ }
	\caption{IDCL: Deep Image Clustering Based on Curriculum Learning and Density Information}
	\label{Algorithm2}
\end{algorithm}

\section{Experiments} \label{Experimental Settings}


\textbf{DataSets.} We utilize ten benchmark datasets, including five grayscale image datasets (MNIST \cite{MNIST}, 
FASHION MNIST\cite{FASHION}, 
EMNIST Letter A-J\cite{Letter}, 
DIGITS\footnote[4]{https://archive.ics.uci.edu/ml/machine-learning-databases/optdigits/}, 
USPS\cite{USPS}), four challenging real-world image datasets (
GTSRB\cite{GTSRB}
, YTF\cite{YTF}
, CIFAR-10\cite{CIFAR-10}
and STL-10\cite{STL-10}
), and one text-based dataset, REUTERS-10K\cite{REUTERS}.

\noindent\textbf{Evaluation Metrics.} We use two standard metrics: clustering accuracy (ACC) and Normalized Mutual Information (NMI). The higher the values [0,1] of these metrics, the better the performance.

\noindent\textbf{Experimental Setup.} We construct our model using a linear block architecture M-512-512-3,072-10 for the Encoder and a symmetric linear block for the Decoder. The number of patches in Figure 1 is consistently set to 16, yielding a patch embedding size of $I_c*\frac{I_h}{4}*\frac{I_w}{4}$. For training specifics, we employ the Adam optimizer \cite{Adam} with a learning rate of 0.01, 100 pre-training epochs, 200 training epochs, a fixed batch size of 256, and the number of clusters $K$ determined by the categories in the corresponding dataset. In terms of hyperparameter settings, we consistently set $\alpha$ to 0.1 and establish the $\lambda_1$ related to $d_c$ and the threshold $\lambda_2$ of density core at 2\% and 5\%, respectively. In our robust training scheme, the initial proportion $\zeta^0$ of the training set is designated as 60\%, whereas the upper limit $\zeta^{max}$ is established at 95\% for all datasets under consideration.


\begin{table}
	\caption{Dataset statistics.}
	\label{Table1}
	\renewcommand\tabcolsep{0.1pt}
	\renewcommand\arraystretch{1}
	   \small
	\centering
	\begin{tabular}{ccccc}
		\toprule
		Dataset    & \# Samples & \# image  & \# classes   & \# type\\
		\midrule
		MNIST &    70000      &      1*28*28        &    10  & Large-Scale\\
		FASHION    &    70000      &      1*28*28        &    10 & Large-Scale\\
		LETTER A-J    &    56000     &      1*28*28        &    10 & Large-Scale\\
		DIGITS    &    1797       &      1*8*8          &    10 & Small-scale\\
		USPS      &    9298       &      1*16*16        &    10 & grayscale image\\
		GTSRB      &    8790       &      3*32*32        &    5 & Complex Contextual\\
		YTF    &    20000      &      3*64*64        &    80 & Complex Contextual, Multi-Cluster\\
		CIFAR-10 &    60000      &      3*32*32        &    10 & Complex Contextual, Large-Scale\\
		STL-10 &    13000      &      3*96*96        &    10 & Complex Contextual\\
		REUTERS-10K &    10000      &      12000        &    4 & Text dataset\\
		\bottomrule   
	\end{tabular}
\end{table}




\vspace{0.5cm}
\subsection{Quantitative Experiments}

\begin{table*}
	\caption{Clustering performance compared with the baseline and state-of-the-art approaches on ten datasets.} 
\label{Table2}
\renewcommand\tabcolsep{3.5pt}
\renewcommand\arraystretch{1.15}
\small
\centerline{
	\begin{tabular}{ccccccccccc}
		\toprule
		Method & CIFAR-10 & STL-10 & GTSRB & YTF & REUTERS-10K & MNIST & LETTER A-J & FASHION & USPS & DIGITS \\
		\midrule
		DEC\cite{DEC}        & 30.1(25.7) & 35.9(27.6) & 65.7(55.1) & 71.5(88.1) & 73.7(49.7) & 86.5(83.7) & 53.1(51.5) & 51.8(54.6) & 76.2(76.7) & 72.3(69.7) \\
		IDEC \cite{IDEC}       & 31.6(27.3) & 37.8(32.4) & 69.9(60.8) & 70.0(87.0) & 75.6(49.8) & 88.1(86.7) & 54.1(52.1) & 52.9(55.7) & 76.1(78.5) & 76.4(71.6)\\
		Deepcluster \cite{Deepcluster}& 37.2*(-) & 69.9*(-) & -(-) & -(-) & 43.1*(-) & 79.7*(66.1*) & -(-) & 54.2*(51.0*) & 56.2*(54.0*) & -(-) \\
		ASPC-DA \cite{ASPC-DA}    & 32.3(27.5) & 41.2(33.8) & 71.5(62.4) & 79.2(89.9) & 77.1(\textbf{63.2}) & 89.9(83.5) & 56.2(53.1) & 60.0(63.3) & 75.3(76.6) & 81.7(74.0) \\
		ClusterGAN \cite{ClusterGAN} & 41.2*(32.3*) & 42.3*(32.3*) & -(-) & -(-) & -(-) & 96.4*(92.1*) & -(-) & -(-) & \textbf{97.0}*(\textbf{93.1}*) & -(-) \\
		VaGAN-SMM \cite{VaGAN} & 30.2*(17.2*) & -(-) & -(-) & -(-) & 82.0*(60.7*) & 95.8*(90.1*) & -(-) & \textbf{67.0}*(67.1*) & -(-) & -(-)  \\
		CC \cite{CC}         & 79.0*(70.5*) & \textbf{85.0*}(76.4*) & -(-) & -(-) & -(-)  & 88.6*(82.0*) & -(-) & -(-) & 80.6*(74.8*) & -(-)  \\
		DipDECK \cite{DipDECK}    & -(-) & -(-) & 61.99(49.5) & 79.7(92.6) & -(-) & 96.1(90.3) & 52.5(\textbf{59.2}) & 63.5(67.5) & 89.1(85.5) & \textbf{88.3}(\textbf{83.2})\\
		LNSCC \cite{LNSCC}      & \textbf{82.0*}(\textbf{71.3*}) & 73.8*(66.2*) & -(-) & -(-) & 81.2*(60.5*) & \textbf{98.3}*(\textbf{97.1}*) & -(-) & 66.5*(\textbf{70.3}*) & \textbf{98.1}*(\textbf{96.9}*) & -(-)  \\
		DCSPC \cite{DCSPC} & -(-) & -(-) & -(-) & -(-) & -(-) & 91.4*(88.2*) & -(-) & 62.9*(65.8*) & 79.1*(82.6*) & -(-) \\
		HC-MGAN \cite{HC-MGAN}    & -(-) & -(-) & -(-) & -(-) & -(-) & 94.3*(90.5*) & -(-) & \textbf{72.1}*(\textbf{69.1}*) & -(-) & -(-) \\
		EDESC \cite{EDESC}      & 62.7*(46.4*) & 74.5*(68.7*) & \textbf{85.0}(\textbf{67.3}) & \textbf{85.5}(\textbf{93.9}) & 82.5*(61.1*)  & 91.3(86.2) & 49.9(43.2) & 63.1*(67.0*) & -(-) & 84.0(79.5) \\
		DeepDPM \cite{DeepDPM}   & -(-) & \textbf{85.0}(\textbf{79.0}) & 79.9(66.9) & 82.1(93.0) & \textbf{83.0}(61.0) & 98.0(94.0) & \textbf{61.2}(58.3) & 62.0(68.0) & 89.0(88.0) & 85.4(77.9) \\
		DMICC \cite{DMICC} & \textbf{82.8*}(\textbf{74.0*}) & 80.0*(68.9*) & -(-) & -(-) & -(-) & -(-) & -(-) & -(-) & -(-) & -(-) \\
		IDCL(Ours)        & 50.7(48.6) & \textbf{85.1}(\textbf{77.9}) & \textbf{90.0}(\textbf{78.9}) & \textbf{95.6}(\textbf{97.5}) & \textbf{83.2}(\textbf{65.3}) & \textbf{98.5}(\textbf{95.9}) & \textbf{73.4}(\textbf{69.6}) & 65.1(63.3) & 95.2(89.5) & \textbf{95.3}(\textbf{91.4}) \\
		\bottomrule
	\end{tabular}
}
\begin{tablenotes}   
	\footnotesize             
	\item[] \textit {All results of baseline algorithms are reported by running their released code with the exception of those denoted by an asterisk (*), which are extracted directly from the respective papers. The symbol "-" indicates that the result is unavailable in either the paper or the code.}      
\end{tablenotes}         
\end{table*}

The quantitative results can be seen in \cref{Table2}, where the Top-2 performances are marked in bold. Additional details of the 14 state-of-the-art clustering approaches can be found in our supplementary material. 

\cref{Table2} reports that our method achieves top-1 performance on 6 out of 10 datasets. More precisely, our model is capable of handling extreme distributions like "multi-cluster," "small-scale," and "high-dimensional." For example, the YTF has a total of 80 clusters and suffers from data imbalance with a maximum difference of 3.85 times, while there are only 1797 samples in DIGITS, which is significantly less than the amount of data required for the general depth model to converge. As a result, our method significantly prevails over these state-of-the-art methods by a large margin, surpassing the second place by 10.1\% and 7\% in ACC, respectively. This is mostly attributable to the cluster-oriented density core in cluster assignment, which enables the model to fully capture the structure of each cluster regardless of cluster size or number.

In addition, our IDCL also performs well on some datasets with complex contexts. In particular, STL-10 is acquired from real environments where the images have diverse backgrounds and are commonly multi-scale or multi-objective; GTSRB records 5 types of traffic signs in different weather conditions; the characters in the LETTER A-J dataset are more complex than MNIST in both shape and case. Our model has good robustness and achieves the best clustering of the above dataset with 0.851, 0.900, and 0.734, respectively, on ACC.

It can also be found from \cref{Table2} that our algorithm excels the similar algorithms ASPC-DA and ClusterGAN on all data sets. Particularly for MNIST, we surpassed ASPC-DA and ClusterGAN by 12.4\% and 3.8\% in terms of NMI\, respectively. We speculate this discrepancy might be because existing counterparts are based on scoring samples by unreliable loss and updating the training set from the whole dataset using a rigid pace. However, it is generally agreed that unsupervised network losses are less stable and reliable than under supervision in most cases due to no ground truth information.

Although our method does not rank first on the CIFAR-10(fourth place), and USPS(third place) datasets, the difference between it and the state-of-the-art is not statistically significant. More importantly, we use fixed parameter values in all datasets without elaborate parameter tuning.

\subsection{Convergence Analysis}
\begin{figure}
	\centering
	\subfloat[Visualization of latent representations at the 0th iteration]
	{\includegraphics[width=0.48\linewidth]{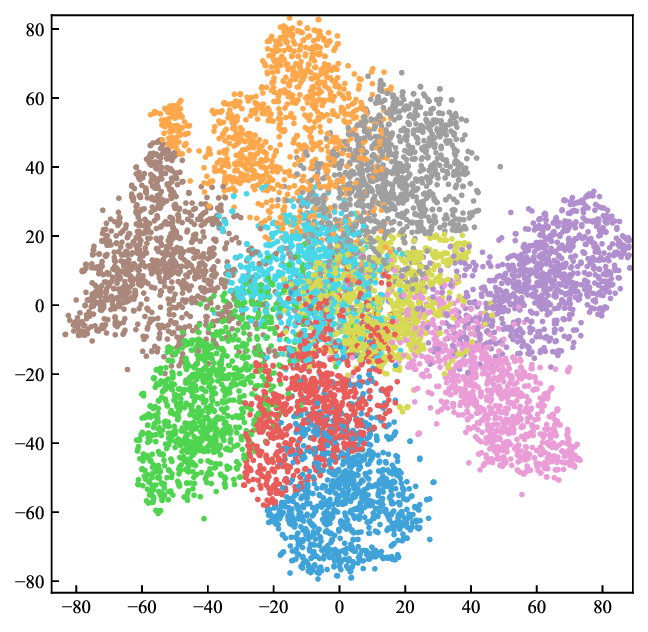}} \hspace{0mm}
	\subfloat[Visualization of latent representations at the 200th iteration]
	{\includegraphics[width=0.48\linewidth]{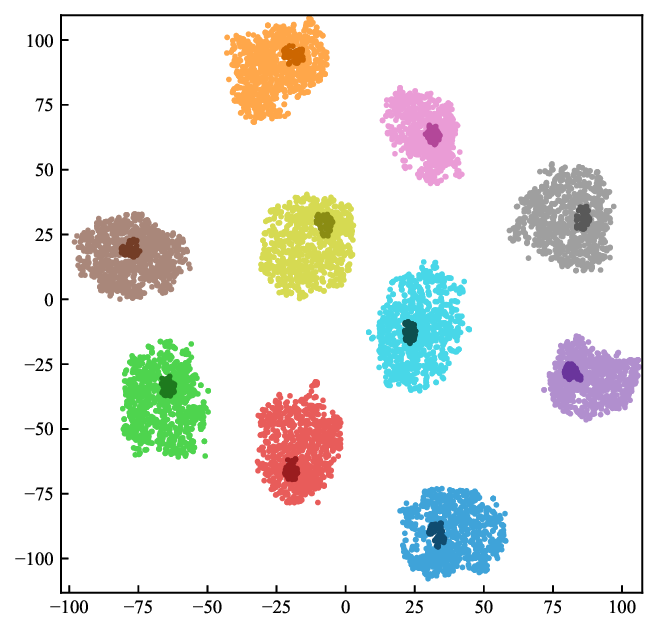}} \vspace{-3mm} 
	\subfloat[Performance vs. iterations]
	{\includegraphics[width=0.8\linewidth]{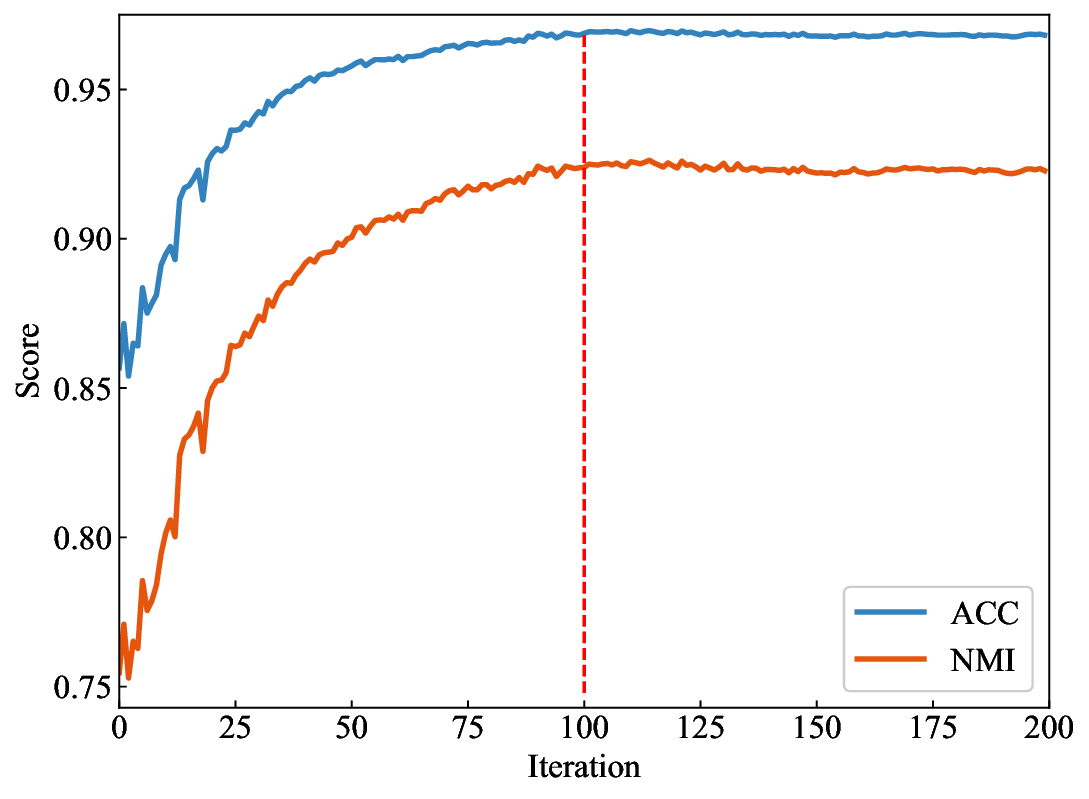}}  \vspace{-3mm}
	\subfloat[Training losses vs. iterations]
	{\includegraphics[width=0.8\linewidth]{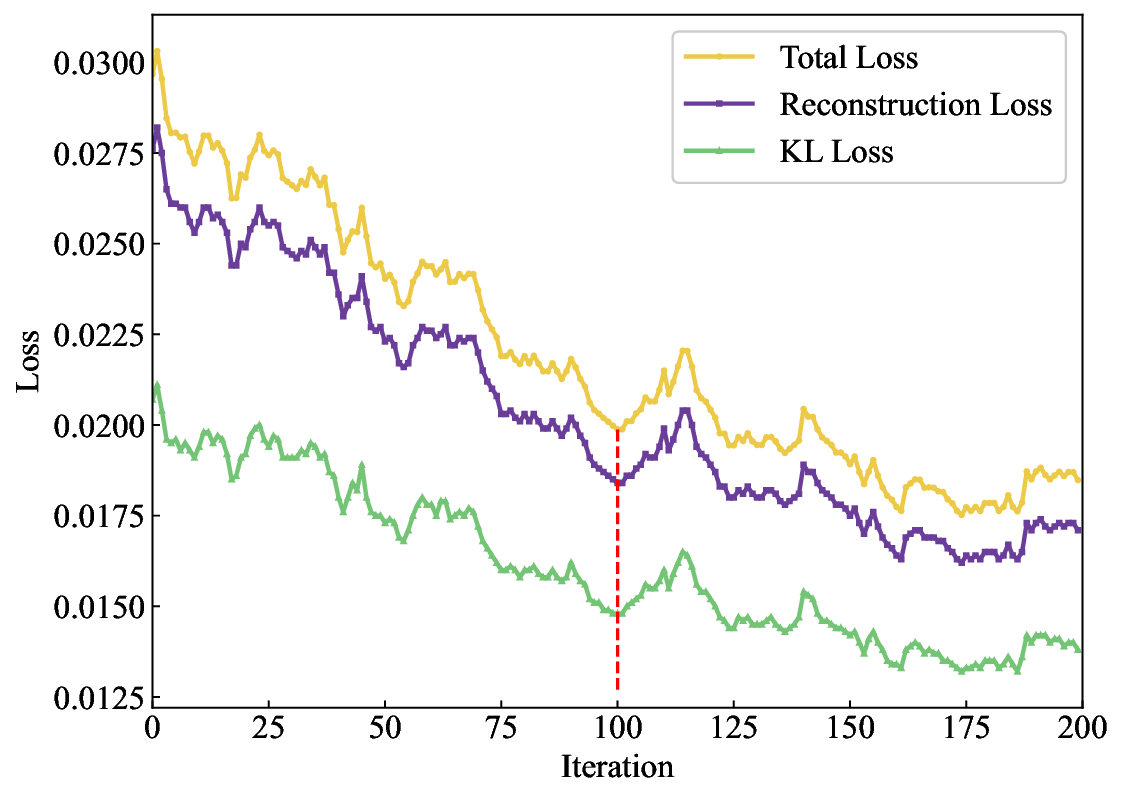}} 
	\caption{The convergence process on MNIST-test dataset.}
	\vspace{-0.25cm}
	\label{Fig3}
\end{figure}

To validate the convergence of the proposed method, we run 200 epochs on the MNIST-test dataset and then visualize the convergence process in \cref{Fig3}. The result in \cref{Fig3}(a) shows that raw data by TSNE are all mixed and unguided, with several clusters squeezed together, implying the difficulty of the clustering. As the training proceeds, the formed clusters become more reasonable, and most data points in the same cluster crowd together after 200 epochs, as shown in \cref{Fig3}(b). In addition, the dark-colored points in \cref{Fig3}(b) represent the density core of each cluster. In \cref{Fig3}(c) and (d), we observe the representation of line graphs that delineate the fluctuations in clustering metrics and loss values as the number of training iterations increases. It is evident that the clustering performance experiences a substantial enhancement during the initial iterations, ultimately attaining a stable state. As depicted in both figures, our model achieves convergence at the 100th iteration, signifying the efficacy of the learning process.


%

\subsection{Running Time Comparison}

%
\cref{Table3} reports the convergence times of IDCL, using the criterion of achieving 90\% of the maximum accuracy. Specifically, our model is better than these state-of-the-art methods. Moreover, the distinct time difference between our model and the degenerate model (lack of training scheme) validates that the proposed course learning scheme plays an essential role in the fast convergence of the model.

Interestingly, our framework is actually faster than the comparable approach ASPC-DA using self-step learning. We believe this is because the proposed scheme uses a more flexible pace, allowing the model to learn in a "slow then fast" manner rather than the traditional fixed pace.

\begin{table}[H]
	\caption{Running time (second). IDCL(w/o CL) denotes our model but does not use the proposed training scheme.}
	\label{Table3}
	\renewcommand\tabcolsep{9pt}
	\renewcommand\arraystretch{1.1}
	   \small
	\centering
	\begin{tabular}{cccc}
		\toprule
		Dataset    & MNIST & STL-10  & LETTER A-J  \\
		\midrule
		DEC &    351.4      &      266.5        &    323.7 \\
		IDEC &    385.4      &      337.9        &    363.1 \\
		EDESC     &    407.5       &      312.4          &    383.5 \\
		\midrule
		ASPC-DA       &    366.9       &      267.8        &    344.2  \\
		\midrule
		IDCL(w/o CL)      &    399.2       &      345.4        &    387.3\\
		IDCL  &    343.6      &      251.2        &    324.7 \\
		\bottomrule
	\end{tabular}
\end{table}

\subsection{Ablation Study}

\begin{figure}[h]
	\centering
	\subfloat[the raw images]
	{\includegraphics[width=0.4\linewidth]{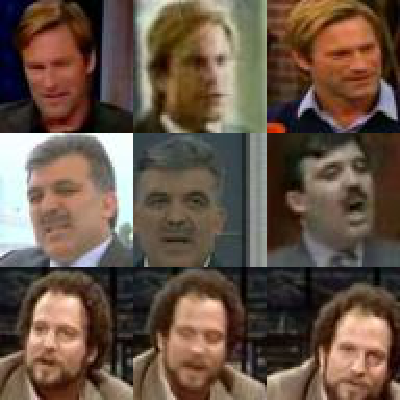}} \hspace{0mm}
	\subfloat[the reconstructed images]
	{\includegraphics[width=0.4\linewidth]{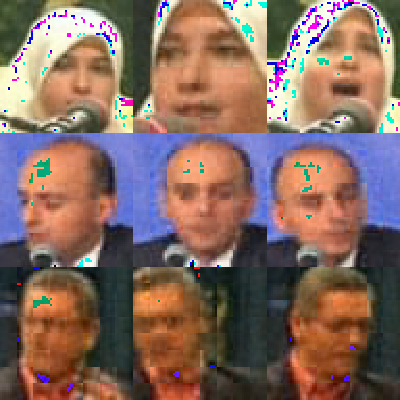}} \vspace{0mm} 
	\caption{Output results of the Decoder on YTF dataset.}   
	\label{Fig5}
\end{figure}

We construct five degradation models for ablation study by removing the corresponding terms. \cref{Table4} summarizes ablation results, from which we can draw some conclusions: 

1) Model-1 is the simplest technique without CL, density core, and Transformer, and its performance is definitely the worst; 2) Model-4 suffers a significant decrease in metrics across all datasets compared to our IDCL, which reveals that the proposed robust training scheme with Curriculum Learning is the most crucial part of our algorithm; 3) the Transformer embodies the great feature representations of complex images since the accuracy drop of Model-3 on YTF (with various backgrounds and multiple clusters) is significantly more than that of other data sets. Likewise, \cref{Fig5} demonstrates the effective reconstruction of the original images by the Decoder when applied to the YTF dataset; 4) the density core can help the model capture the inherent cluster structure and enhance the supervised signals, which is especially friendly for small-scale data scenarios that cannot be adequately trained, such as the DIGITS dataset listed in \cref{Table4}.

\vspace{.5cm}
\begin{table}[h]
	\caption{Ablation study results of the proposed method and its degradation models.}
	\label{Table4}
	\renewcommand\tabcolsep{3pt}
	\renewcommand\arraystretch{1.15}
	\footnotesize
	\centering
	\begin{threeparttable}
		\begin{tabular}{ccccccc}
			\toprule[1pt]
			Methods	& $\mathbf C_k^{\rm core}$ & TB & RCL & MNIST   & DIGITS   & YTF   \\
			\midrule[0.5pt]
			Model-1   & $\times$ & $\times$ & $\times$ & \begin{tabular}[c]{@{}c@{}}ACC:0.893\\ NMI:0.872\end{tabular} & \begin{tabular}[c]{@{}c@{}}ACC:0.775\\ NMI:0.724\end{tabular} & \begin{tabular}[c]{@{}c@{}}ACC:0.701\\ NMI:0.852\end{tabular} \\
			\midrule[0.5pt]
			Model-2  & $\times$  & \checkmark  & \checkmark & \begin{tabular}[c]{@{}c@{}}ACC:0.961\\ NMI:0.935\end{tabular} & \begin{tabular}[c]{@{}c@{}}ACC:0.899\\ NMI:0.835\end{tabular} & \begin{tabular}[c]{@{}c@{}}ACC:0.928\\ NMI:0.945\end{tabular} \\
			\midrule[0.5pt]
			Model-3    & \checkmark & $\times$ & \checkmark & \begin{tabular}[c]{@{}c@{}}ACC:0.970\\ NMI:0.937\end{tabular} & \begin{tabular}[c]{@{}c@{}}ACC:0.926\\ NMI:0.887\end{tabular} & \begin{tabular}[c]{@{}c@{}}ACC:0.864\\ NMI:0.901\end{tabular} \\
			\midrule[0.5pt]
			Model-4    & \checkmark & \checkmark & $\times$ & \begin{tabular}[c]{@{}c@{}}ACC:0.913\\ NMI:0.922\end{tabular} & \begin{tabular}[c]{@{}c@{}}ACC:0.937\\ NMI:0.894\end{tabular} & \begin{tabular}[c]{@{}c@{}}ACC:0.912\\ NMI:0.963\end{tabular} \\
			\midrule[1pt]
			\begin{tabular}[c]{@{}c@{}}Model-5\\ (Fixed pace)\end{tabular}    & \checkmark & \checkmark & \checkmark & \begin{tabular}[c]{@{}c@{}}ACC:0.974\\ NMI:0.950\end{tabular} & \begin{tabular}[c]{@{}c@{}}ACC:0.932\\ NMI:0.901\end{tabular} & \begin{tabular}[c]{@{}c@{}}ACC:0.932\\ NMI:0.969\end{tabular} \\
			\midrule[0.5pt]
			Ours & \checkmark & \checkmark & \checkmark & \begin{tabular}[c]{@{}c@{}}ACC:\textbf{0.985}\\ NMI:\textbf{0.958}\end{tabular} & \begin{tabular}[c]{@{}c@{}}ACC:\textbf{0.953}\\ NMI:\textbf{0.914}\end{tabular} & \begin{tabular}[c]{@{}c@{}}ACC:\textbf{0.956}\\ NMI:\textbf{0.975}\end{tabular} \\
			\bottomrule[1pt]
		\end{tabular}
		\begin{tablenotes}   
			\footnotesize             
			\item[] \textit {"$\mathbf C_k^{\rm core}$" denotes the Density Core, "TB" represents the Transformer block, "RCL" denotes our robust curriculum Learning-based scheme}      
		\end{tablenotes}           
	\end{threeparttable} 
\end{table}
\vspace{-.5cm}

\section{Conclusion}

In this paper, we propose a robust deep image clustering based on curriculum learning and density information (IDCL), which is capable of handling image data with challenging scenarios such as multiple-cluster, small-scale, large-scale, and complex backgrounds. Specifically, the density information is first employed to objectively evaluate the difficulty of the samples. Subsequently, the principles of "easy first" for the sample and "slow first" for the learning pace provide the model with dynamically incremental training sets in each training iteration. With this in mind, the robust training scheme based on Curriculum Learning enables robust and efficient model learning. Also, the proposed density kernel replaces the widely used but unstable cluster centers during cluster assignment, thereby generating a more reasonable supervisory signal. The quantitative experiment, ablation study, convergence analysis, and running time comparison further demonstrate the effectiveness of IDCL.

\bibliographystyle{ACM-Reference-Format}
\balance
\bibliography{main}

\clearpage
\onecolumn
\appendix

\section{Supplementary Material}
In this supplementary material, we supplement the derivation process of $L_{clu}$ in IDCL. 
Second, we present the details of the datasets used in our experiments, the processing of each dataset, and the data enhancement approach (Section~\ref{app:sec-datasets}). 
Then, we describe the data sources for the comparison algorithms (Section~\ref{app:sec-algorithms}). 
Finally, we provide a detailed explanation of the computational procedure for the metrics used in the experiments (Section~\ref{app:sec-metrics}).

\section{Derivation of the Gradient of $L_{clu}$ w.r.t. the Embedded Point $z_i$ in IDCL}
\label{app:sec-derivation}

In the main text, we define the following formula:
\begin{equation}
\label{app:eq-grad-main}
\begin{aligned}
\frac{\partial L_{clu}}{\partial z_i} 
= 2\sum_{k=1}^K 
\left(
\frac{p_{ik}-q_{ik}}
{\sum_{x_j \in \mathbf C_k^{\rm core}} (1+\|z_{i}-z_j\|^2)^{-1}}
\right)
\cdot
\left(
\sum_{x_j \in \mathbf C_k^{\rm core}} 
\frac{\|z_i-z_j\|}
{(1+\|z_i-z_j\|^2)^2}
\right).
\end{aligned}
\end{equation}

The derivation of the above equation is as follows.

Given:
\begin{equation}
\label{app:eq-lclu}
L_{clu}=\sum_{i=1}^{|\mathcal D|} \sum_{k=1}^{K} p_{ik} \log \frac{p_{ik}}{q_{ik}},
\end{equation}

\begin{equation}
\label{app:eq-qik}
q_{ik} = \frac{\sum_{x_j \in \mathbf C_k^{\rm core}} (1+\|z_i-z_j\|^2)^{-1}}
{\sum_{k=1}^K \sum_{z_j \in \mathbf C_k^{\rm core}} (1+\|z_i-z_j\|^2)^{-1}},
\end{equation}

\begin{equation}
\label{app:eq-pik}
p_{ik} = \dfrac{q_{ik}^2 / \sum_{i=1}^{|\mathcal D|} q_{ik}}
{\sum_{k=1}^{K} \left(q_{ik}^2 / \sum_{i=1}^{|\mathcal D|} q_{ik}\right)}.
\end{equation}

Calculate: $\frac{\partial L_{clu}}{\partial z_i}$.

For convenience, define
\begin{equation}
\label{app:eq-dik}
d_{ik}=\sum_{x_j \in \mathbf C_k^{\rm core}} (1+\|z_i-z_j\|^2)^{-1},
\end{equation}

\begin{equation}
\label{app:eq-ai}
A_i=\sum_{k=1}^K \sum_{z_j \in \mathbf C_k^{\rm core}} (1+\|z_i-z_j\|^2)^{-1}
=\sum_{k=1}^K d_{ik}.
\end{equation}

Then \cref{app:eq-qik} can be simplified as:
\begin{equation}
\label{app:eq-qik-simple}
q_{ik}=\frac{d_{ik}}{A_i}.
\end{equation}

When $i$ varies, for all $k$, $d_{ik}$ will change. According to the chain rule,
\begin{equation}
\label{app:eq-chain}
\begin{aligned}
\frac{\partial L_{clu}}{\partial z_i}
&= \sum_{k=1}^K \frac{\partial L_{clu}}{\partial d_{ik}} \cdot \frac{\partial d_{ik}}{\partial z_i}\\
&= \sum_{k=1}^K \frac{\partial L_{clu}}{\partial d_{ik}}
\cdot
\frac{\partial \sum_{x_j \in \mathbf C_k^{\rm core}} (1+\|z_i-z_j\|^2)^{-1}}{\partial z_i}\\
&= -2\sum_{k=1}^K \frac{\partial L_{clu}}{\partial d_{ik}}
\cdot
\left(
\sum_{x_j \in \mathbf C_k^{\rm core}} 
\frac{\|z_i-z_j\|}{(1+\|z_i-z_j\|^2)^2}
\right).
\end{aligned}
\end{equation}

Then calculate $\frac{\partial L_{clu}}{\partial d_{ik}}$.  
When $d_{ik}$ changes, according to \cref{app:eq-ai}, $A_i$ changes, and according to \cref{app:eq-qik-simple}, $q_{i1},q_{i2},\cdots,q_{iK}$ change. However, for all $q_{lm}$ with $l\neq i$, they remain unchanged. Thus,
\begin{equation}
\label{app:eq-dldd}
\begin{aligned}
\frac{\partial L_{clu}}{\partial d_{ik}}
&= \frac{\partial \left(\sum_l \sum_m p_{lm}\log \frac{p_{lm}}{q_{lm}}\right)}{\partial d_{ik}}\\
&= \sum_l \sum_m \frac{\partial(p_{lm}\log p_{lm}-p_{lm}\log q_{lm})}{\partial d_{ik}}\\
&= -\sum_l \sum_m \frac{\partial(p_{lm}\log q_{lm})}{\partial d_{ik}}\\
&= -\sum_m \frac{\partial(p_{im}\log q_{im})}{\partial d_{ik}}.
\end{aligned}
\end{equation}

Substituting \cref{app:eq-qik-simple} into the above equation, we obtain
\begin{equation}
\label{app:eq-dldd-final}
\begin{aligned}
\frac{\partial L_{clu}}{\partial d_{ik}}
&= -\sum_m p_{im} \frac{\partial(\log(d_{im}/A_i))}{\partial d_{ik}}\\
&= -\sum_m p_{im} \frac{\partial(\log(d_{im}))}{\partial d_{ik}}
+\sum_m p_{im} \frac{\partial(\log(A_i))}{\partial d_{ik}}\\
&= -p_{ik}\frac{\partial(\log(d_{ik}))}{\partial d_{ik}}
+\frac{1}{A_i}\cdot \frac{\partial A_i}{\partial d_{ik}}\cdot \sum_m p_{im}\\
&= -\frac{p_{ik}}{d_{ik}} + \frac{1}{A_i}\cdot \frac{\partial \sum_t d_{it}}{\partial d_{ik}}\\
&= -\frac{p_{ik}}{d_{ik}} + \frac{1}{A_i}\\
&= \frac{q_{ik}-p_{ik}}{d_{ik}}.
\end{aligned}
\end{equation}

In the above calculation, from the third line to the fourth line, the property $\sum_m p_{im}=1$ is used, which can be directly obtained from \cref{app:eq-pik}. Substituting \cref{app:eq-dldd-final} and \cref{app:eq-dik} into \cref{app:eq-chain}, we obtain
\begin{equation}
\label{app:eq-grad-final}
\begin{aligned}
\frac{\partial L_{clu}}{\partial z_i}
=
2\sum_{k=1}^K 
\left(
\frac{p_{ik}-q_{ik}}
{\sum_{x_j \in \mathbf C_k^{\rm core}} (1+\|z_i-z_j\|^2)^{-1}}
\right)
\cdot
\left(
\sum_{x_j \in \mathbf C_k^{\rm core}} 
\frac{\|z_i-z_j\|}
{(1+\|z_i-z_j\|^2)^2}
\right).
\end{aligned}
\end{equation}

Note: the symbols used in this appendix are intended solely for mathematical derivation and may differ from those used in the algorithm.

\section{Comparison Algorithms}
\label{app:sec-algorithms}

We compared our proposed approach to a range of state-of-the-art algorithms, including Deepcluster\cite{Deepcluster}, ClusterGAN\cite{ClusterGAN}, VaGAN-SMM\cite{VaGAN-SMM}, CC\cite{CC}, LNSCC\cite{LNSCC}, DCSPC\cite{DCSPC}, HC-MGAN\cite{HC-MGAN}, and DMICC\cite{DMICC}. The results for these algorithms were taken from the original or related papers. The results for the ASPC-DA\cite{ASPC-DA} algorithm were obtained through our own implementation, using an AE structure of [500, 500, 1000, 10, 1000, 500, 500], a batch size of 256, 500 rounds of pre-training, and the ADAM optimizer for both pre-training and formal training. The k-means algorithm used in this experiment was also set to its default setting.

The DEC\cite{DEC} algorithm used source \href{https://github.com/XifengGuo/DEC-keras}{code-DEC} as its foundation, and the experimental setup included an AE structure of [500, 500, 2000, 10, 2000, 500, 500], a batch size of 256, and 500 pre-training rounds. The ADAM optimizer and a learning rate of 0.01 were used throughout pre-training and formal training. The IDEC\cite{IDEC} algorithm, which used source \href{https://github.com/XifengGuo/IDEC}{code-IDEC}, had the same experimental setup as DEC. For the DipDECK\cite{DipDECK} algorithm, which was based on source \href{https://dmm.dbs.ifi.lmu.de/cms/downloads/index.html}{code-DipDECK}, the initial number of clusters was set to the correct number, and all other parameters used the default settings in the source code. Similarly, the DeepDPM\cite{DeepDPM} algorithm that used source \href{https://github.com/BGU-CS-VIL/DeepDPM}{code-DeepDPM} had a setup similar to the previous DipDECK, with the exception of the initial number of clusters being set to the correct number and all other parameters using the default settings from the source code. The results for the IDCEC\cite{IDCEC} algorithm were taken from the original article. Lastly, the EDESC\cite{EDESC} algorithm, which was based on source \href{https://github.com/JinyuCai95/EDESC-pytorch}{code-EDESC}, utilized the default configuration in the source code. A brief introduction of the comparison algorithms is shown in \cref{app:tab-algorithms}.

\begin{table}[H]
\centering
\caption{Comparison algorithms statistics.}
\label{app:tab-algorithms}
\renewcommand\tabcolsep{25pt}
\renewcommand\arraystretch{1}
\begin{tabular}{ccc}
\toprule
Name & \# source & \# data source \\
\midrule
DEC\cite{DEC} & 2016/ICML & our implementation \\
IDEC\cite{IDEC} & 2017/IJCAI & our implementation \\
DeepCluster\cite{Deepcluster} & 2018/ECCV & original or related papers \\
ASPC-DA\cite{ASPC-DA} & 2020/TKDE & our implementation \\
ClusterGAN\cite{ClusterGAN} & 2019/CVPR & original or related papers \\
VaGAN-SMM\cite{VaGAN-SMM} & 2022/TNNLS & original or related papers \\
CC\cite{CC} & 2021/AAAI & original or related papers \\
DipDECK\cite{DipDECK} & 2021/SIGKDD & our implementation \\
LNSCC\cite{LNSCC} & 2022/IJCAI & original or related papers \\
DCSPC\cite{DCSPC} & 2022/--- & original or related papers \\
HC-MGAN\cite{HC-MGAN} & 2022/TNNLS & original or related papers \\
EDESC\cite{EDESC} & 2022/CVPR & our implementation \\
DeepDPM\cite{DeepDPM} & 2022/CVPR & our implementation \\
DMICC\cite{DMICC} & 2023/AAAI & original or related papers \\
\bottomrule
\end{tabular}
\end{table}

\section{Discussion and Future Directions}
\label{subsec:openworld_future}

\noindent \textbf{Towards Open-World Perception.}
The robust learning of latent representations via curriculum pacing is a transformative cornerstone for transitioning from static clustering to the dynamic Open-World paradigm. Under the broader framework of Open-World Semi-Supervised Learning \cite{zheng2024textual, zheng2024prototypical}, a fundamental bottleneck persists: the unrealistic assumption that highly noisy and complex data distributions can be effectively disentangled without a carefully guided learning strategy. By synergizing IDCL with Transformer-based global dependency modeling \cite{zhang2023tdec} and multi-modality co-teaching \cite{zheng2024textual}, future architectures can achieve a holistic, autonomous understanding of evolving data streams, entirely bypassing the need for human-in-the-loop parameter tuning.  The implications of this autonomous capability extend far beyond discovery, naturally bridging the gap to data-efficient learning and generative modeling. Specifically, by dynamically capturing the underlying semantic modes of complex distributions via our "slow-then-fast" pacing, the proposed logic serves as a critical structural prior. This guidance is instrumental for adaptive dataset quantization \cite{LiZDXQ25} and diverse dataset distillation frameworks \cite{libeyond,li2026fixed,li2026efficient}, ensuring that synthetic or compressed proxies faithfully preserve the topological and categorical diversity of the original data. Furthermore, accurately recognizing latent cluster densities allows diverse diffusion-based augmentations \cite{LiZ0XLQ24} to be anchored to precise semantic centers, thereby substantially enhancing data-free knowledge distillation. Finally, in the realm of advanced 3D generative vision, these density-aware structural priors show immense potential for refining probability density flow matching, enabling more stable and mode-preserving geodesic estimations for novel view synthesis \cite{wang2026geodesicnvs}.

\noindent \textbf{Broader Impact and Cross-Domain Versatility.}
By automating the pacing of feature learning and effectively mitigating error accumulation, IDCL systematically dismantles the structural bottlenecks inherent in classical deep clustering. This capability positions our framework as a highly versatile initialization engine, empowering a broad spectrum of downstream pipelines. In the realm of medical image analysis, IDCL facilitates precise pathological region partitioning by progressively disentangling the subtle variances among distinct tissue layers and lesion typologies in OCT scans, providing essential structural priors for guided denoising and segmentation \cite{li2026cross,li2025retidiff,li2024efficient,li2026garnet}. Within 3D vision, our density-aware clustering enhances global solvers and vanishing point estimation \cite{zhao2026advances,edstedt2024dedode,li2023hong,liao2024globalpointer} by extracting robust geometric representations from highly unstructured environments, while naturally streamlining instance-level grouping in complex point clouds \cite{qu2024conditional,qu2025end,qu2025robust}. Transitioning to physical intelligence and robotics, IDCL automates the discovery of latent physical modes—such as slip, rotation, or stable contact—within high-frequency sensor and tactile data streams. By anchoring these modes to stable density cores, it significantly refines stability analysis and multi-agent cooperative control \cite{yan2025pandas,ruan2024q,yan2023stability,zhang2025ccma,xu2025sedm,tian2025measuring}. For Embodied AI and video-language modeling, the dynamic pacing of IDCL drives temporal event partitioning by adaptively clustering video segments into coherent semantic sequences \cite{li2025lion}. Concurrently, it assists in optimizing the granularity of action primitives within Vision-Language-Action (VLA) models \cite{li2025cogvla}, laying the structured semantic groundwork required for higher-order, System-2 reasoning \cite{song2025hume}. Beyond spatio-temporal modeling, our framework optimizes knowledge distillation by extracting high-confidence density cores for location-aware semantic masking \cite{lan2025acam,lan2026clockdistill,lan2024gradient}, and yields aligned latent representations that serve as a robust bridge for semi-supervised domain translation via generative diffusion models \cite{wang2025ladb}. Finally, the scalability of IDCL proves highly effective for open-vocabulary entity grouping in macro-scale domains like marine and remote sensing segmentation \cite{li2025exploring,li2025stitchfusion,li2025maris,li2025exploring2}, as well as for discovering underlying structural patterns in complex socio-meteorological analyses \cite{liu2026health,shen2025aienhanced,shen2026mftformer}.

\section{Clustering Metrics}
\label{app:sec-metrics}

To evaluate the clustering performance, we adopt two standard evaluation metrics: Accuracy (ACC) and Normalized Mutual Information (NMI).

The best mapping between cluster assignments and true labels is computed using the Hungarian algorithm to measure accuracy \cite{Hungarian}. For completeness, we define ACC by:
\begin{equation}
\label{app:eq-acc}
\begin{aligned}
ACC=\max_{m}\frac{\sum_{i=1}^{n}\mathbf{1}\left\{l_i=m(c_i)\right\}}{n},
\end{aligned}
\end{equation}
where $l_i$ and $c_i$ are the ground-truth label and predicted cluster label of data point $x_i$, respectively.

NMI calculates the normalized measure of similarity between two labels of the same data:
\begin{equation}
\label{app:eq-nmi}
\begin{aligned}
NMI=\frac{I(l;c)}{\max\{H(l),H(c)\}},
\end{aligned}
\end{equation}
where $I(l,c)$ denotes the mutual information between true label $l$ and predicted cluster $c$, and $H$ represents their entropy. Results of NMI do not change by permutations of clusters and are normalized to [0,1].

\section{Data Sets}
\label{app:sec-datasets}

The details of the datasets used for the experiments are described below. To ensure appropriate dataset sizing, we cropped the selected images to dimensions of $3 \times 64 \times 64$ pixels.

\noindent
\textbf{CIFAR-10}\cite{CIFAR-10}: The CIFAR-10 dataset consists of 60,000 $32\times32$ color images in 10 classes. There are 6,000 images per class, with 5,000 training and 1,000 testing images per class.\\
\textbf{STL-10}\cite{STL-10}: STL-10 is acquired from real environments where the images have diverse backgrounds and are commonly multi-scale or multi-objective. It contains 10 classes: airplane, bird, car, cat, deer, dog, horse, monkey, ship, and truck.\\
\textbf{MNIST}\cite{MNIST}: Greyscale image dataset consisting of 70,000 handwritten digits (0 to 9) with a size of $28\times28$ pixels. The training set contains 60,000 images and the test set contains 10,000 images.\\
\textbf{Fashion-MNIST}\cite{FMNIST}: Greyscale image dataset consisting of 70,000 articles from the Zalando online store. Each sample belongs to one of 10 product groups and has a size of $28\times28$ pixels.\\
\textbf{USPS}\cite{USPS}: Greyscale image dataset consisting of 9,298 handwritten digits (0 to 9) with a size of $16\times16$ pixels.\\
\textbf{Digits}\cite{Digits}: Greyscale image dataset consisting of 1,797 handwritten digits (0 to 9) with a size of $8\times8$ pixels.\\
\textbf{GTSRB}\cite{GTSRB}: Color image dataset consisting of 8,790 traffic sign images (5 categories) with a size of $3\times32\times32$ pixels. To ensure balance, we chose the five categories with the highest number of samples.\\
\textbf{YTF}\cite{YTF}: Color image dataset consisting of 20,000 face images (80 identities) with a size of $3\times64\times64$ pixels. We selected 80 people in alphabetical order from each category and cropped the images to the proper size.\\
\textbf{Letter}\cite{Letter}: Greyscale image dataset consisting of 56,000 handwritten letters (A to J) with a size of $28\times28$ pixels.\\
\textbf{REUTERS-10K}\cite{REUTERS}: Reuters contains about 810,000 English news stories categorized in a tree structure. Following DEC\cite{DEC}, we focus on four root categories: corporate/industrial, government/social, markets, and economics, excluding documents with multiple labels.

\begin{figure}[H]
\centering
\subfloat[MNIST]{\includegraphics[width=2.5cm]{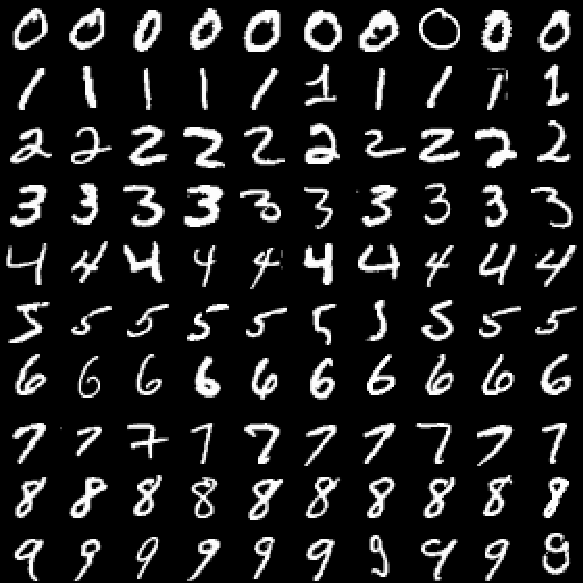}}
\quad
\subfloat[FMNIST]{\includegraphics[width=2.5cm]{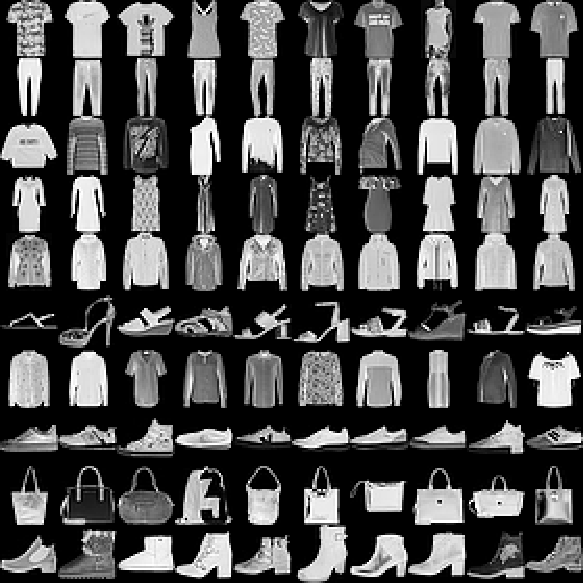}}
\quad
\subfloat[USPS]{\includegraphics[width=2.5cm]{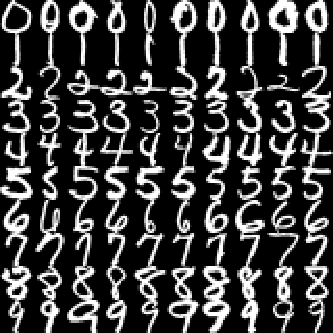}}
\quad
\subfloat[Letter]{\includegraphics[width=2.5cm]{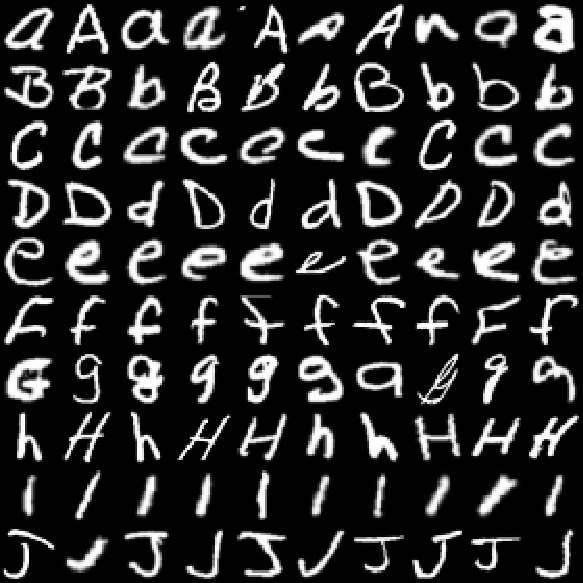}}
\quad
\subfloat[Digits]{\includegraphics[width=2.5cm]{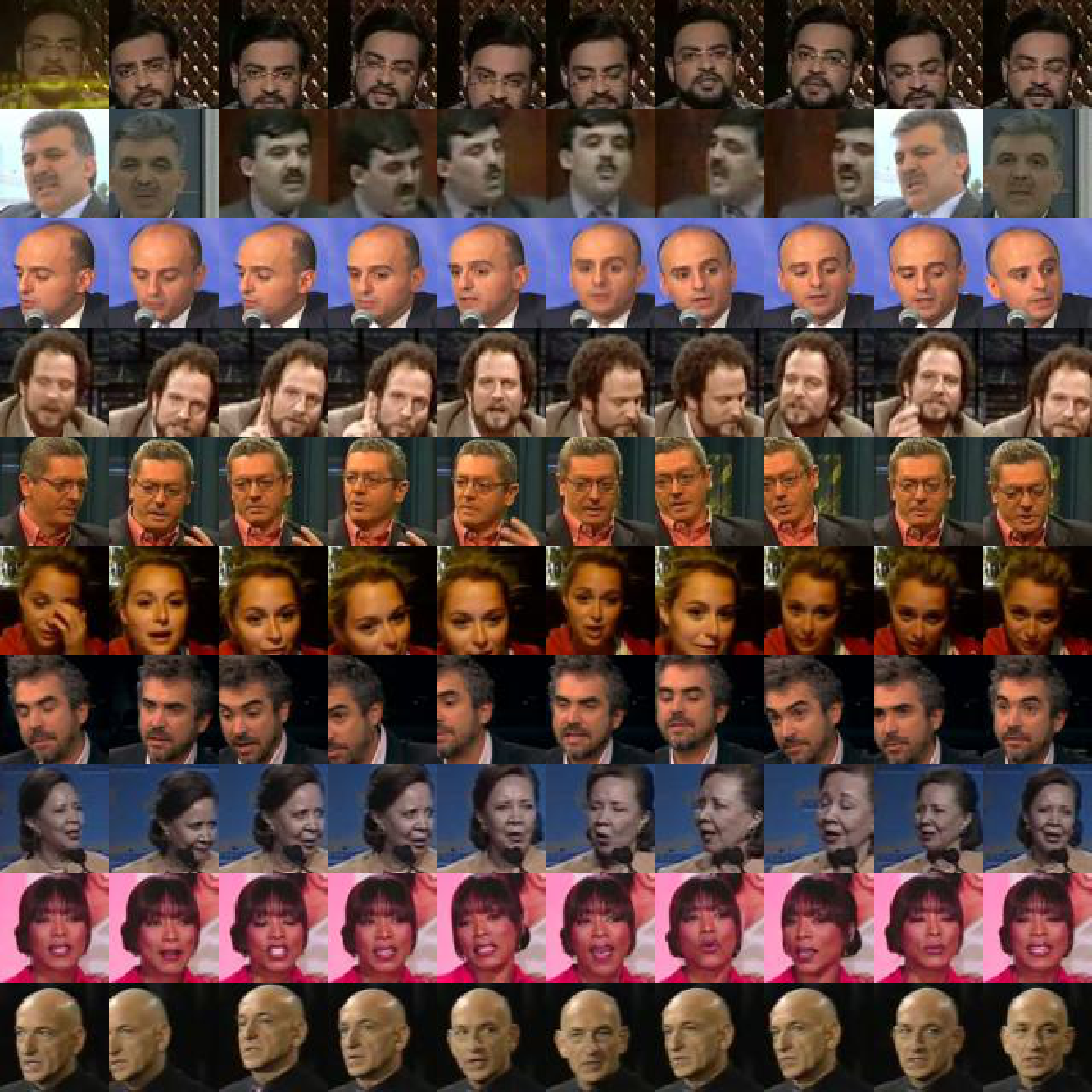}}
\quad
\subfloat[GTSRB]{\includegraphics[width=2.5cm]{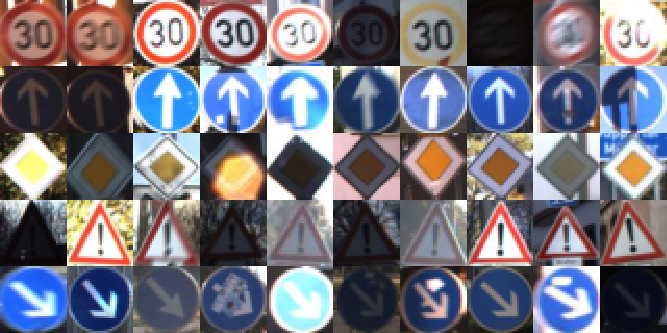}}
\caption{Datasets visualization.}
\label{app:fig-datasets}
\end{figure}

In order to enhance the image data used during model training, we employed two methods: randomly rotating the images up to 10 degrees and shifting them by a maximum of $\frac{H}{10}$ pixels in any direction, where $H$ is the height of the input image. In addition, as the patch size for sinusoidal position encodings must be even, we performed a resize operation on part of the dataset, primarily converting images of size $28\times28$ to $32\times32$ using standard library functions. Examples of all datasets are shown in \cref{app:fig-datasets}.

\end{document}